\documentclass[letterpaper, 10 pt, journal, twoside]{IEEEtran} %

\IEEEoverridecommandlockouts                              %
\usepackage{etex}
\usepackage{graphics} %
\usepackage{epsfig} %
\usepackage{newtxtext}
\usepackage{times} %
\usepackage{amsfonts} 
\usepackage{amsmath} %
\usepackage{amssymb}  %
\usepackage{bm}
\usepackage{float}
\usepackage{pgf}
\usepackage{hhline}
\usepackage{hyperref}
\usepackage{cite}
\usepackage{multirow}
\usepackage{colortbl}
\usepackage{array}
\usepackage{relsize}
\usepackage{xspace}
\usepackage{color}
\usepackage{graphics}
\usepackage{graphicx}

\usepackage{amsmath}

\usepackage{amsfonts}
\usepackage{amsmath}
\usepackage{amssymb}

\usepackage{subcaption}
\usepackage{float}
\usepackage{todonotes}
\usepackage{url} 
\usepackage{multirow}
\usepackage{hhline}
\usepackage[capitalise]{cleveref}
\usepackage{placeins}

\usepackage{booktabs}
\newcommand{\ra}[1]{\renewcommand{\arraystretch}{#1}}

\title{Self-Supervised Domain Calibration and Uncertainty Estimation for Place
  Recognition}%

\author{Pierre-Yves Lajoie, Giovanni Beltrame%
\thanks{ 
  This work was partially supported by a Canadian Space Agency FAST
  Grant and a Vanier Canada Graduate Scholarships Award.
 }%
\thanks{Department\,of\,Computer\,and\,Software\,Engineering, \mbox{Polytechnique Montr\'eal}, Montreal, Canada, \newline
{\tt\scriptsize\,\{pierre-yves.lajoie, giovanni.beltrame\}@polymtl.ca}}
}

\begin{document}

\maketitle

\begin{abstract}
Visual place recognition techniques based on deep learning, which have imposed
themselves as the state-of-the-art in recent years, do not generalize
well to environments visually different from the training set. Thus, to
achieve top performance, it is sometimes necessary to fine-tune the networks to
the target environment.
To this end, we propose a self-supervised
domain calibration procedure based on robust pose graph optimization from
Simultaneous Localization and Mapping (SLAM) as the supervision signal without
requiring GPS or manual labeling. 
Moreover, we leverage the procedure to improve uncertainty estimation for place recognition matches which is important in safety critical applications.
We show that our approach
can improve the performance of a state-of-the-art technique on a target
environment dissimilar from its training set and that we can obtain uncertainty estimates. 
We believe that this approach will
help practitioners to deploy robust place recognition solutions in
real-world applications.
Our code is available publicly: \url{https://github.com/MISTLab/vpr-calibration-and-uncertainty} 
\begin{IEEEkeywords}
Place Recognition, Uncertainty Estimation, \mbox{Simultaneous} Localization And Mapping
\end{IEEEkeywords}
\end{abstract}

\begin{tikzpicture}[overlay, remember picture]
  \path (current page.north east) ++(-4,-0.2) node[below left] {
  This paper has been accepted for publication in the IEEE Robotics and Automation Letters.
  };
  \end{tikzpicture}
  \begin{tikzpicture}[overlay, remember picture]
  \path (current page.north east) ++(-5.5,-0.6) node[below left] {
   Please cite the paper as: P. Lajoie and G. Beltrame,
  };
  \end{tikzpicture}
  \begin{tikzpicture}[overlay, remember picture]
  \path (current page.north east) ++(-4.4,-1) node[below left] {
  ``Self-Supervised Domain Calibration and Uncertainty Estimation for Place Recognition'',
  };
  \end{tikzpicture}
  \begin{tikzpicture}[overlay, remember picture]
  \path (current page.north east) ++(-7.1,-1.4) node[below left] {
    IEEE Robotics and Automation Letters (RA-L), 2023.
  };
  \end{tikzpicture}
\newcommand{\name}[1]{{\smaller \sf#1}\xspace}
\newcommand{\acronym}[1]{{#1}\xspace}
\newcommand{\SLAM}{\acronym{\mbox{SLAM}}}
\newcommand{\CSLAM}{\acronym{\mbox{C-SLAM}}}

\newcommand{\prob}[1]{\textcolor{red}{#1}}
\newcommand{\modif}[1]{\textcolor{green}{#1}}

\newcommand{\norm}[1]{\left\lVert#1\right\rVert}

\newcommand{\sugg}[2]{\textcolor{blue}{\sout{#1}#2}}
\newcommand{\py}[1]{\textcolor{red}{#1}}

\section{Introduction}

\begin{figure*}[h]
    \centering
    \includegraphics[width=\textwidth,trim=0mm 0mm 0mm 0mm,clip]{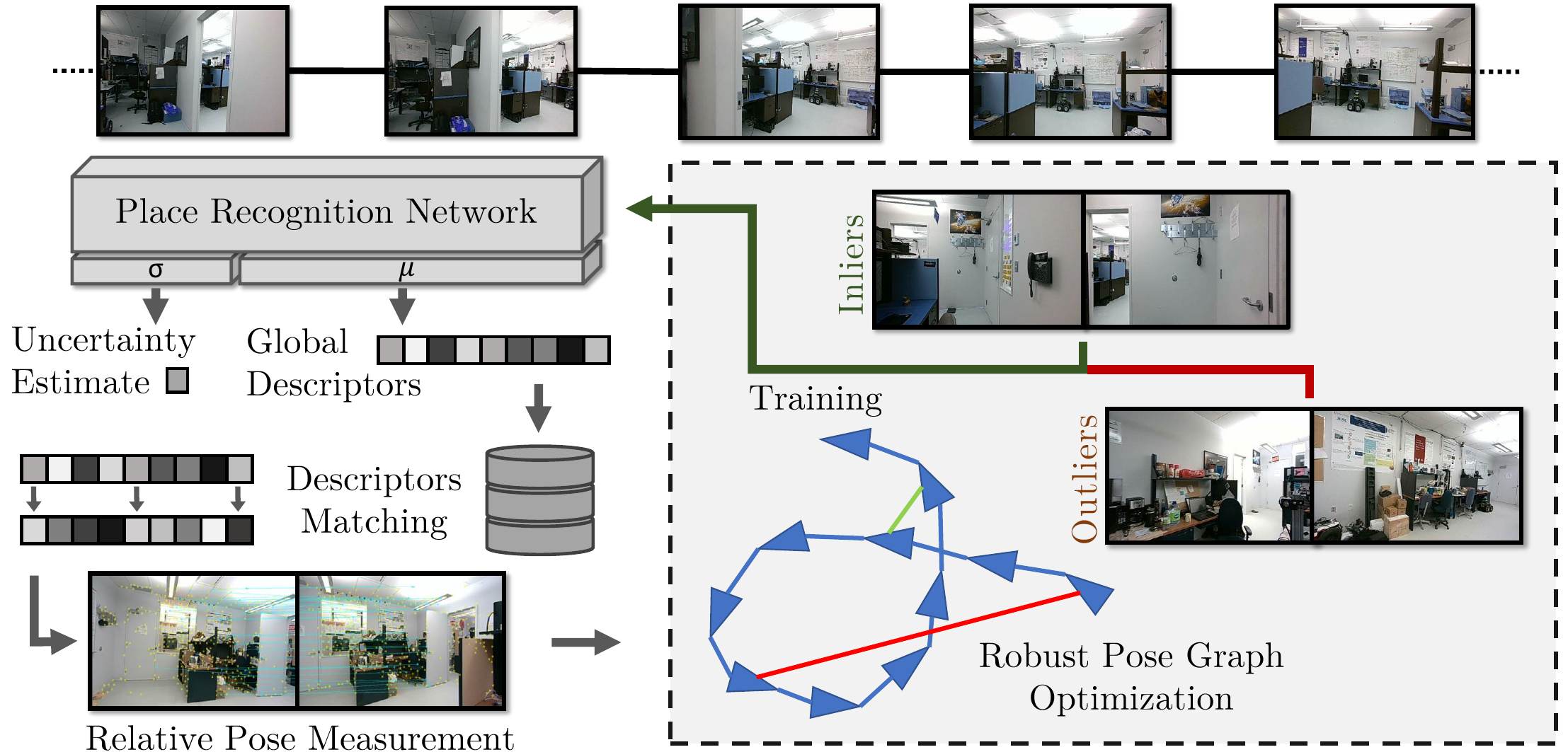}
    \caption{\textbf{Self-Supervised Domain Calibration and Uncertainty
        Estimation via Robust SLAM:} Using a single calibration sequence through
      a new environment, our proposed self-supervised technique for visual place
      recognition verifies putative loop closures using recent progress in
      robust pose graph optimization, and uses both the resulting inliers and
      outliers to fine-tune the place recognition network. The place recognition
      network tuned to the new domain achieves better
      performance on subsequent sequences in visually similar environments, and
      provides uncertainty estimates tailored to those environments. Our
      calibration approach does not rely on GPS or any ground truth information,
      and can thus improve place recognition systems in any environment. }
    \label{fig:top-level}
    \vspace{-5mm}
\end{figure*}

\IEEEPARstart{V}{isual} Place Recognition (VPR) remains one of the core problems of
autonomous driving and long-term robot localization. Recognizing
previously visited places is essential for decision-making, to reduce
localization drift in Simultaneous Localization and Mapping (SLAM), and to
improve robots' situational awareness in
general~\cite{lowryVisualPlaceRecognition2016}. While VPR techniques based on
deep learning can achieve very high levels of accuracy on standard
datasets~\cite{barrosPlaceRecognitionSurvey2021}, domain generalization is still
a major concern when the deployment environment is visually and/or structurally
different from the training data. The
problem of domain discrepancies is especially important for indoors or
subterranean deployments since most popular approaches are trained using GPS
data on city streets
images~\cite{arandjelovicNetVLADCNNArchitecture2018,hauslerPatchNetVLADMultiScaleFusion2021,leyva-vallinaGeneralizedContrastiveOptimization2021}.

Generalization and feature transferability from one domain to another, are
common issues in deep learning~\cite{yosinskiHowTransferableAre2014}.
The most common and effective approach is still to calibrate or fine-tune,
the representation to the testing domain. Given that most robots are deployed in
known domains (e.g. roads, warehouses, etc.), one can refine the network using
additional labeled samples directly from the known testing environment to tailor
the representation to the target domain. While an effective approach, the data
labelling necessary to obtain new training samples can be prohibitively
expensive in practice. 

To solve this problem, we believe that robust SLAM can be used as a self-supervised tool for data mining in any environment without the need for external sensors or ground truth information. 
In standard SLAM, place recognition errors are known to cause catastrophic localization
failures. However, recent
progress in robust state estimation has shown that such erroneous VPR matches
can be detected and removed during pose graph optimization
(PGO)~\cite{lajoieModelingPerceptualAliasing2019,yangGraduatedNonConvexityRobust2020}.
In other words, robust PGO leverages the 3D structure of the environment and robot trajectory to classify VPR matches as correct and incorrect.
Both correct and incorrect matches can in turn be used to fine-tune VPR networks
to improve their performance or obtain uncertainty estimates.

Therefore, in this paper, we propose a self-supervised
domain calibration approach to extract new training samples from any target domains
and improve VPR networks accuracy. In
addition, we propose a technique to train an uncertainty estimator for place
recognition using the new extracted samples.

First, we show that our self-supervised approach to gather training samples can
be used to train a VPR network from a pretrained classification model and
achieve reasonable performance, thus demonstrating the strength of our
self-supervised training signal. We then show that our approach can
improve the performance of existing VPR solutions when applied to environments
that are visually different from their training domain, as well as providing uncertainty estimates.

Previous self-supervised
approaches~\cite{arandjelovicNetVLADCNNArchitecture2018,pillaiSelfSupervisedVisualPlace2017}
relied on GPS localization to extract training samples from datasets by
selecting images with minimal distance. However, this is not suitable
for GPS-denied environments such as indoor, underwater or underground. 
Also, contrary to techniques
using structure-from-motion (SfM) for data
mining~\cite{radenovicFineTuningCNNImage2019}, our approach leverages additional
outlier samples identified with robust SLAM to further improve the VPR network.
Moreover, our approach is able to extract samples in any environment in which
odometry estimates can be obtained, leveraging sensors such as IMUs and
wheel encoders that are not used in SfM.

Our approach offers practical benefits for the deployment of VPR systems in
real applications: it could be used to collect correct and incorrect training
samples from a single calibration run through an environment similar to the
target domain, or could be employed online for lifelong learning/tuning directly
on the target environment. 
After calibration, the
VPR network is able to detect more correct matches and identify uncertain
images. Moreover, by producing fewer incorrect matches, it reduces the expensive
computational burden of processing and rejecting
them~\cite{carsonPredictingImproveIntegrity2022}.
Our contributions can be summarized as follows:
\begin{itemize}
\item A self-supervised training samples extraction method that does not
  require any external sensor (e.g. GPS), ground truth or manual labelling;
\item A VPR sample classification method in correct and incorrect matches based on robust SLAM estimates;
\item A domain calibration procedure for existing VPR techniques to improve their
  performance on any target environment using both correct and incorrect samples.
\item An uncertainty estimator leveraging the new correct and incorrect samples during training;
\item Open-source packages for sample extraction, network refinement and uncertainty estimation.
\end{itemize}

In the rest of this paper, \cref{sec:related} presents some background knowledge
and related work, \cref{sec:method} details the proposed approach,
\cref{sec:exp} demonstrate the effectiveness of the technique, and
\cref{sec:conclusion} offers conclusions and discusses future work.
\section{Background and Related Work}
\label{sec:related}

\subsection{Visual Place Recognition}

The ability to recognize places is crucial for
localization, navigation, and augmented reality, among other
applications~\cite{gargWhereYourPlace2021}. The most popular approach is to
compute and store global descriptors for each image to match, followed by an
image retrieval scheme using a database of descriptors. Global descriptors are
usually represented as high-dimensionality vectors which can be compared
with simple distance functions (e.g. Euclidean or cosine distance) to obtain a similarity
metric between two images. The seminal work of
NetVLAD~\cite{arandjelovicNetVLADCNNArchitecture2018} extracts descriptors using
a CNN and leverages Vectors of Locally Aggregated
Descriptors~\cite{jegouAggregatingLocalDescriptors2010} to get a representation
well-suited for image retrieval. The descriptor network is typically trained
using tuples of images mined from large datasets. An anchor image is first
chosen, then positive and negative samples are selected based on close and far
GPS localization, respectively. A triplet margin loss pushes the network to
output similar representations for positive and anchor samples and dissimilar
representations for negative ones. Recent work has extended the concept of
global descriptors by extracting local-global descriptors from patches in the
feature space of each image~\cite{hauslerPatchNetVLADMultiScaleFusion2021}. In
another line of work, \cite{leyva-vallinaGeneralizedContrastiveOptimization2021}
proposed a Generalized Contrastive loss (GCL) function that relies on image
similarity as a continuous measure instead of binary labels (i.e., positive and
negative samples).

Recent works in place recognition have aimed at computing uncertainty estimates
for individual samples (i.e., images) using an uncertainty-aware loss during
training~\cite{tahaUnsupervisedDataUncertainty2019,warburgBayesianTripletLoss2021b,caiSTUNSelfTeachingUncertainty2022a}.
This loss function allows the system to reduce its confidence and the priority
of samples with high uncertainty. Similar to standard place recognition, the
uncertainty estimates are dependent on the training domain, meaning it is
beneficial to train those estimates on the target domain. In this work, we use
the Bayesian Triplet Loss from \cite{warburgBayesianTripletLoss2021b} for
correct samples and a Kullback-Leibler divergence loss for incorrect samples as
described in \cref{subsec:uncertainty}.

\subsection{Robust SLAM}

In SLAM, place recognition is used to produce loop closure measurements between
the current pose (i.e., rotation and translation) of a robot and the
pose corresponding to the last time it has visited the place. Loop closure
measurements are combined with odometry (i.e., egomotion) measurements in a
graph representing the robot/camera trajectory. In other words, the SLAM
algorithm builds a pose graph with odometry links between subsequent poses and
loop closure links between recognized places. Pose graph optimization is then
performed to reduce the localization drift of the
robot~\cite{cadenaPresentFutureSimultaneous2016}. When using a global descriptor
method, such as NetVLAD, VPR serves as a first filter through potential matches,
which is followed by the more expensive task of feature matching and
registration to obtain the relative pose measurement corresponding to the loop
closure. Due to the occurrence of perceptual aliasing (i.e., when two distinct
similar-looking places are confused as the same), some loop closure measurements
are incorrect and, if left undetected, they can lead to dramatic localization
failures~\cite{lajoieModelingPerceptualAliasing2019}. This phenomenon is
particularly important when computing loop closures between multiple robots maps
for collaborative
localization~\cite{lajoieCollaborativeSimultaneousLocalization2022}.

To mitigate the negative effect of incorrect loop closure measurements during
pose graph optimization, several approaches have been proposed. They vary from
adding decision variables to the optimization
problem~\cite{lajoieModelingPerceptualAliasing2019},
to leveraging clusters in the pose graph
structure~\cite{wuClusterbasedPenaltyScaling2020}.
For the purpose of this paper, we chose a recent approach based on Graduated
Non-Convexity which as been shown to efficiently achieve superior
results~\cite{yangGraduatedNonConvexityRobust2020}.

It is important to note that these approaches allow us indirectly to classify
loop closure measurements, and by extension also VPR matches, as correct
(inliers) or incorrect (outliers).

\subsection{Domain Calibration}

The goal of domain calibration is to improve the performance of a system on a
target domain that is different from the training domain. This can be done
through fine-tuning the model using samples from the target domain or
through more complex domain adaptation approaches to enhance the generalization
ability of the model. 
Domain calibration is also a major concern for long-term visual localization in changing environments. 
As presented in~\cite{churchillExperiencebasedNavigationLongterm2013}, one approach is to store multiple maps of the same environment to account for scene variation.
To ensure the scalability, \cite{muhlfellnerSummaryMapsLifelong2016} proposes to summarize the maps, 
and \cite{doanScalablePlaceRecognition2019,doanHM4HiddenMarkov2020} suggest the use of compressed or coarse representions based on Hidden Markov Models.

Our approach could be use to adapt the VPR network to appearance changes. 
In fact, approaches in that line of research have been proposed for sequence
adaption to cope with changing weather conditions during long-term
missions~\cite{poravDonWorryWeather2019,schubertGraphbasedNonlinearLeast2021}.

Interestingly, the exploitation of local feature patterns has been identified as
a key to domain adaptation since they are more generic and transferable than
global approaches~\cite{wenExploitingLocalFeature2019}. Alternatively, recent
work have proposed to include geometric and semantic information into the VPR
latent embedding representation for visual place
recognition~\cite{huDASGILDomainAdaptation2021} to better adapt to the target
domain. In another line of work, \cite{chenSelfSupervisedVisualPlace2022} uses
temporal and feature neighborhoods in panoramic sensor data to mine training
samples for VPR fine-tuning: they classify samples as correct using
geometric verification, as opposed to our work where we leverage recent
progress in robust SLAM.

Unlike related techniques based on GPS data~\cite{pillaiSelfSupervisedVisualPlace2017},
our approach is suited to any environment in which an odometry system (i.e.,
visual inertial odometry, lidar-based odometry paired with
VPR, etc.) can be deployed, such as
indoors, subterranean, or underwater. 
Our approach also requires significantly less data than SfM-based approaches~\cite{radenovicFineTuningCNNImage2019}.
Moreover, we include incorrect matches
corresponding to loop closing outliers in the learning process to avoid
such occurrences in the improved network. In addition, by adding an uncertainty
head to the VPR network and using a Bayesian Triplet
Loss~\cite{warburgBayesianTripletLoss2021b}, we are able to train an estimator
for the heteroscedastic aleatoric
uncertainty~\cite{kiureghianAleatoryEpistemicDoes2009} (i.e, uncertainty
corresponding to a particular data input) using only the extracted samples.
\section{Self-Supervised Domain Calibration and Uncertainty Estimation}
\label{sec:method}

The main challenge addressed by our method is to extract pseudo-ground truth
labels for training images. To tune the model to the target domain, we need to
gather positive and negative place recognition matches from a single preliminary
run through the environment. The classic approach is to use external positioning
systems (e.g. GPS) to identify images that where captured in the same location
as positive samples and images captured in distant locations as negative
samples~\cite{arandjelovicNetVLADCNNArchitecture2018,pillaiSelfSupervisedVisualPlace2017}.
We aim to extend this data mining scheme to any environment, regardless of the
availability of ground truth localization. Our process is split in three
sequential steps. First, we perform SLAM on an initial run through the target
environment with a camera, or robot. In particular, we compute visual odometry,
and we gather putative VPR matches from an initial network that was not tuned to
the specific environment. Second, we sort the putative matches as correct or
incorrect samples using robust pose graph optimization. Third, we use all the
resulting samples to fine-tune the VPR network to the target domain and train an
uncertainty estimator. A summary of the method is illustrated
in~\cref{fig:top-level}. 

\subsection{Finding VPR Matches}

Global images descriptors can be
complex structures such as Vector of Locally Aggregated
Descriptors~\cite{jegouAggregatingLocalDescriptors2010}, used in
NetVLAD~\cite{arandjelovicNetVLADCNNArchitecture2018}, or simply the features
extracted from the penultimate layer of a standard classification
network~\cite{zhouPlacesImageDatabase2017}. The descriptors are represented as a
vector $f(I_i)$, where $f$ is the image representation extraction function and
$I_i$ the $i_{th}$ keyframe.
As keyframes are processed, we store the computed global descriptors in a
database. Then, for each keyframe we query the best matches using nearest
neighbors search, by sorting the global descriptors based on the Euclidean
distance $d(q,I_i)$ between the query descriptor $f(q)$ and the other images
descriptors $f(I_i)$. This results in a sorted list of the best putative VPR
matches for each keyframe for the run through the environment. To avoid trivial
matches in the same location, we do not consider matches with keyframes in the vicinity of the query.

\subsection{Classifying Matches}

To filter the VPR matches, we first compute the relative pose between the pairs
of images and integrate this information, as loop closures, in the SLAM pose
graph. For each keyframe, we compute the relative pose between itself and the
first image in its associated list (the best match). If we are able to
successfully compute a relative pose measurement (i.e., loop closure), we store
the two images as a (anchor, positive) training sample. Otherwise, we repeat the
process with the next best match in the list. To obtain the negative samples, we
go through the remaining best VPR matches in the sorted list and select up to
$N$ images for which a loop closure cannot be computed due to a
lack of keypoint correspondences. $N$ is set according to the available GPU memory for training.
 This way, we ensure that we extract the
negative samples that appear the most similar to the anchor, yet that are not
sufficiently similar to compute a loop closure. In other words, we select the
most valuable negative samples for training, since they represent invalid VPR
matches made by the uncalibrated network. This results in training tuples (1
anchor, 1 positive, $N$ negatives) for each keyframe in the sequence.

\begin{figure}[h]
    \centering
    \begin{subfigure}[b]{0.49\columnwidth}
        \includegraphics[width=\columnwidth,trim=0mm 0mm 0mm 0mm,clip]{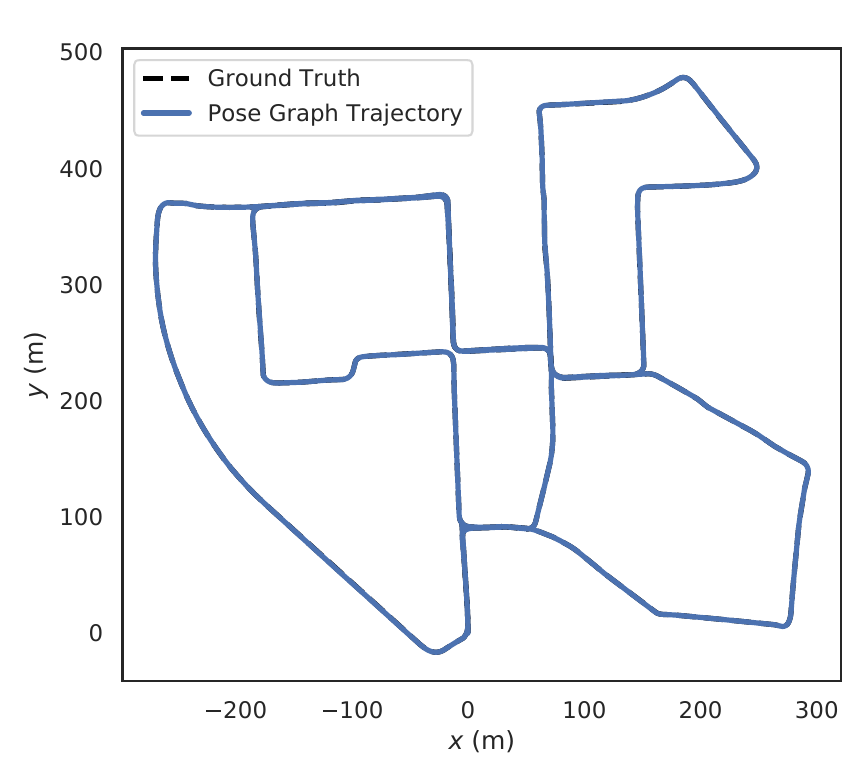}
        \caption{With correct loop closures}
    \end{subfigure}
    \begin{subfigure}[b]{0.49\columnwidth}
        \includegraphics[width=\columnwidth,trim=0mm 0mm 0mm 0mm,clip]{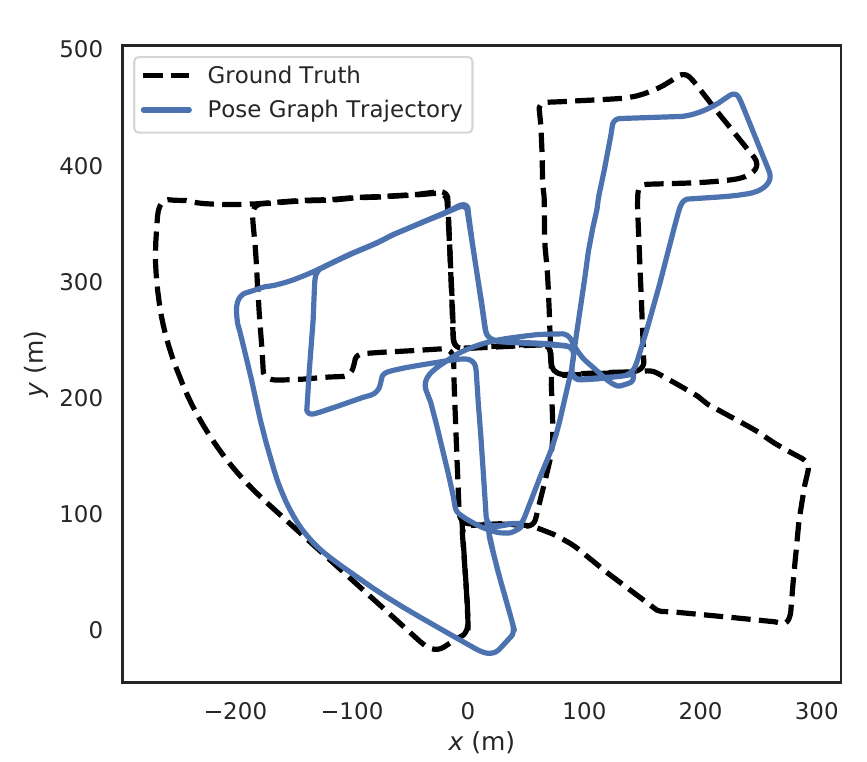}
        \caption{With 1 incorrect loop closure}
    \end{subfigure}
    \caption{Illustration of the resulting KITTI00 pose graphs with and without an incorrect loop closure. We can see the large negative effect of even a single incorrect measurement on the localization accuracy. This motivates the need to detect such incorrect VPR matches using robust pose graph optimization. Those incorrect matches correspond to some of the most confusing parts of the environment and can thus be used in training to futher improve VPR networks.}
    \label{fig:pgo}
\end{figure}

Given the possible occurrence of perceptual aliasing, the computability of a
relative pose measurement between the current anchor and positive frames, is not enough
to guarantee that it is a correct place recognition match (see \cref{fig:pgo}).
Thus, we add the computed relative pose measurements
to the SLAM pose graph as a loop closure and perform robust estimation using the
Graduated Non-Convexity method~\cite{yangGraduatedNonConvexityRobust2020}.

From the resulting optimized pose graph, we can compute the error associated
with every measurement and classify the VPR matches as correct or incorrect. A
large error means that the match is in contradiction with the geometric
structure of the pose graph and therefore incorrect. We then sort the
measurements into the subsets of training samples $S_{correct}$ and
$S_{incorrect}$. The two subsets will be used with different loss functions
during training.

\subsection{Domain Calibration}
\label{subsec:domain-calib}

The domain calibration of our VPR network is done through fine-tuning using the
filtered training tuples in the tuning sets. In other words, starting from the
pretrained network, we performed additional training iterations using the
extracted data. 

For the subset of correct samples, we applied the triplet margin loss $L$ for
each training tuple $(q,p^q,\{n^q_i\}) \in S_{correct}$,
\begin{align}
    L &= \sum_i \max(d(q,p^q) + m - d(q, n^q_i), 0)
\end{align}
where $m$ is the margin, $q$ is the global descriptor of the query image, $p^q$
is the global descriptor of the positive image associated with the query, and
$n^q_i$ are the corresponding negative samples descriptors. The global
descriptors are $1 \times K$ vectors resulting from a forward pass through the
VPR network and the distance function $d$ is the Euclidean distance between the
vectors. This strategy is analog to the training method used in
NetVLAD~\cite{arandjelovicNetVLADCNNArchitecture2018}.

On the other hand, the incorrect samples $S_{incorrect}$ 
are composed of only
one query and one negative images $(q, n^q)$ and do not contain a positive image $p$ such that we cannot use the triplet margin loss.
Therefore, for each incorrect
sample, we use a negative \textit{Mean Squared Error} loss to increase the
distance between the corresponding descriptors,
\begin{align}
    L &= - \frac{1}{K} \sum_k^K (q_k - n^q_k)^2
\end{align}
At each epoch we train on both correct and incorrect samples.

\begin{figure*}[h!]
    \centering
    \begin{subfigure}[b]{0.30\textwidth}
        \includegraphics[width=\textwidth,trim=0mm 0mm 0mm 0mm,clip]{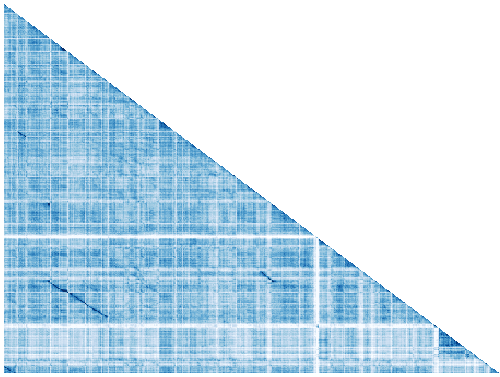}
        \caption{Initial}
    \end{subfigure}
    \begin{subfigure}[b]{0.30\textwidth}
        \includegraphics[width=\textwidth,trim=0mm 0mm 0mm 0mm,clip]{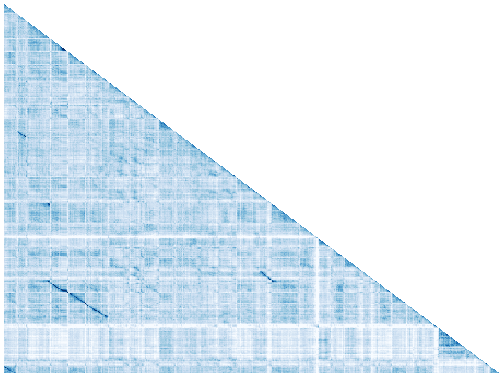}
        \caption{Tuned}
    \end{subfigure}
    \begin{subfigure}[b]{0.30\textwidth}
        \includegraphics[width=\textwidth,trim=0mm 0mm 0mm 0mm,clip]{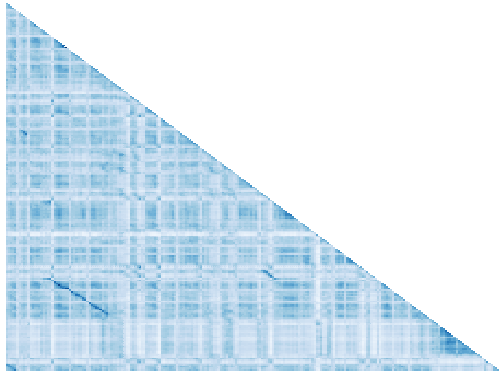}
        \caption{NetVLAD}
    \end{subfigure}
    \begin{subfigure}[b]{0.037\textwidth}
        \includegraphics[width=\textwidth,trim=0mm 0mm 0mm 0mm,clip]{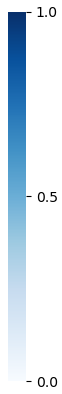}
        \caption*{}
    \end{subfigure}
    \caption{Self-Supervised learning of a visual-similarity metric. An
      illustration of the similarity matrix before (Initial) and after (Tuned)
      training compared with the similarity obtained from NetVLAD on the
      KITTI-00 sequence. As expected, the similarity between positive pairs has
      increased (blue), and it has decreased between negative pairs (white).}
    \label{fig:distance_matrix_training_kitti}
\end{figure*}

\begin{figure*}[h!]
    \centering
    \begin{subfigure}[b]{0.245\textwidth}
        \includegraphics[width=\textwidth,trim=0mm 0mm 0mm 0mm,clip]{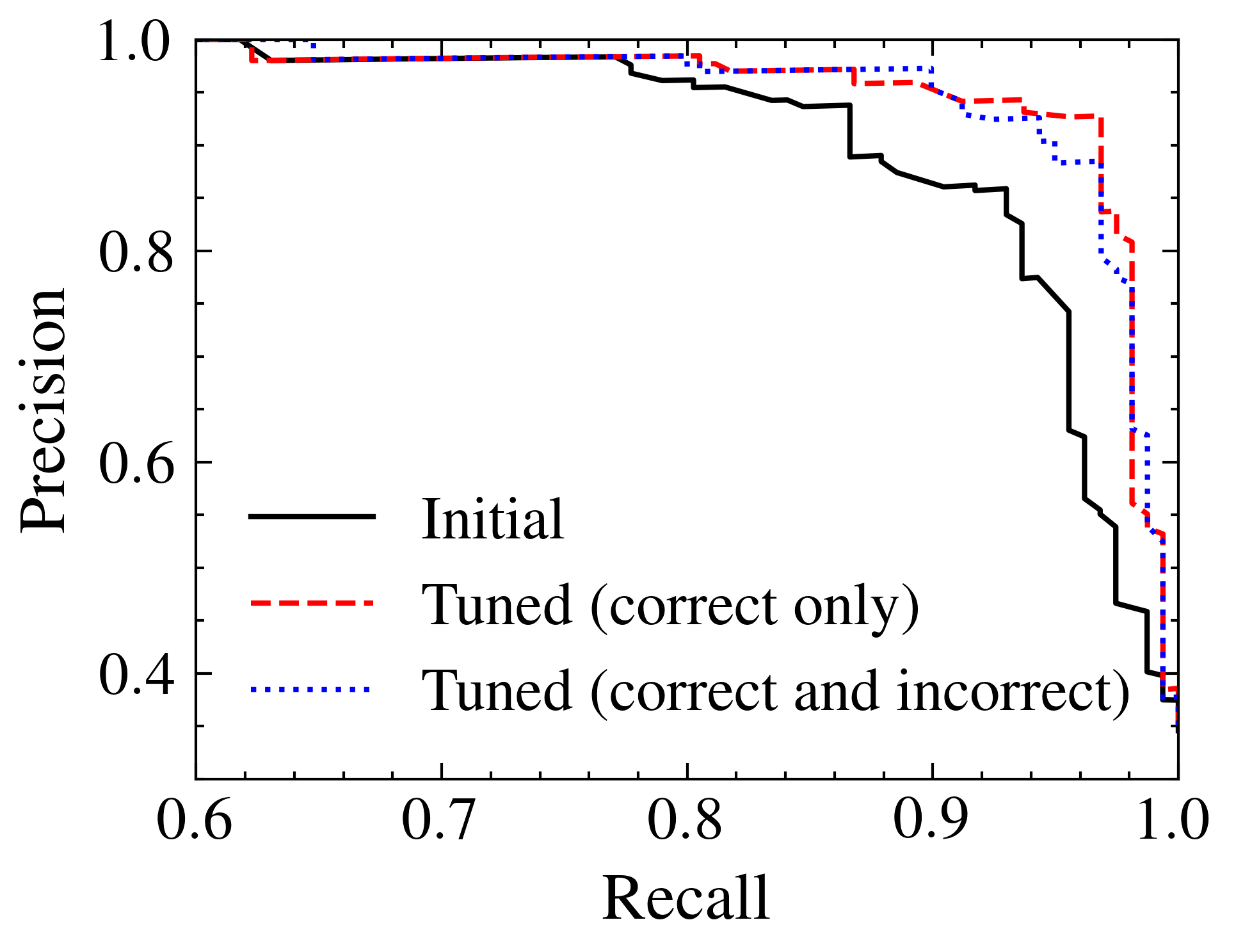}
        \caption{NetVLAD KITTI~00}
    \end{subfigure}
    \begin{subfigure}[b]{0.245\textwidth}
        \includegraphics[width=\textwidth,trim=0mm 0mm 0mm 0mm,clip]{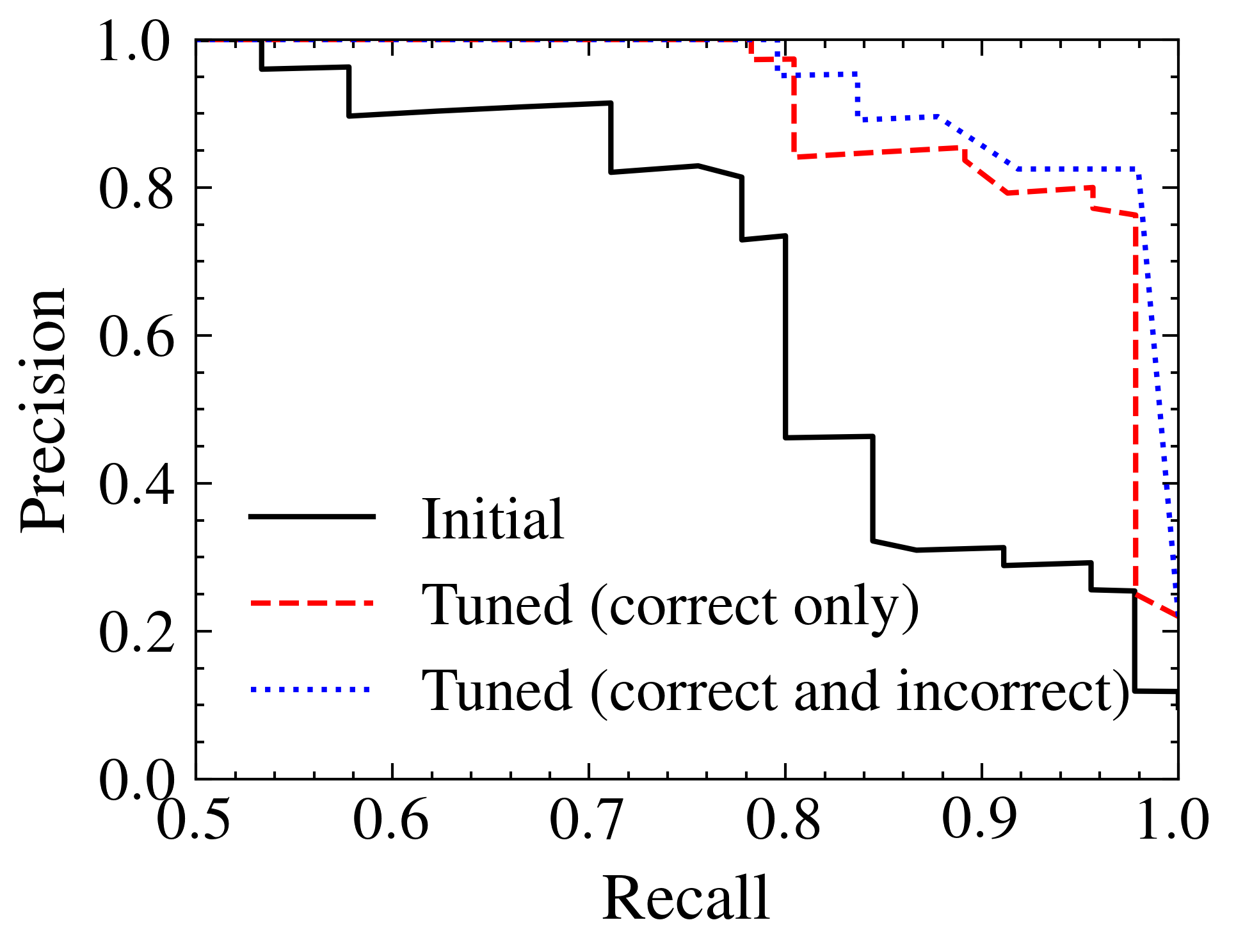}
        \caption{NetVLAD KITTI~02}
    \end{subfigure}
    \begin{subfigure}[b]{0.245\textwidth}
        \includegraphics[width=\textwidth,trim=0mm 0mm 0mm 0mm,clip]{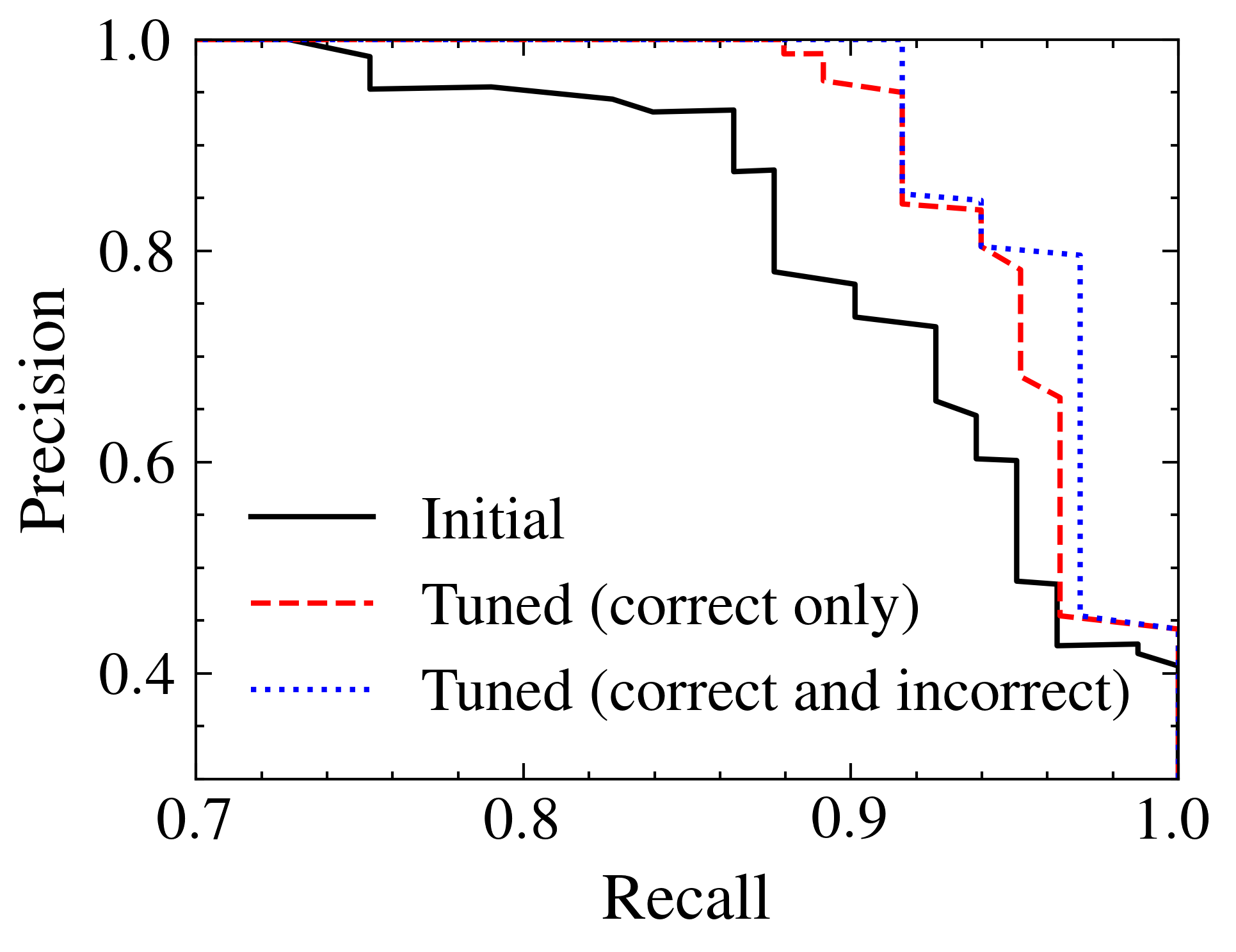}
        \caption{NetVLAD KITTI~05}
    \end{subfigure}
    \begin{subfigure}[b]{0.245\textwidth}
        \includegraphics[width=\textwidth,trim=0mm 0mm 0mm 0mm,clip]{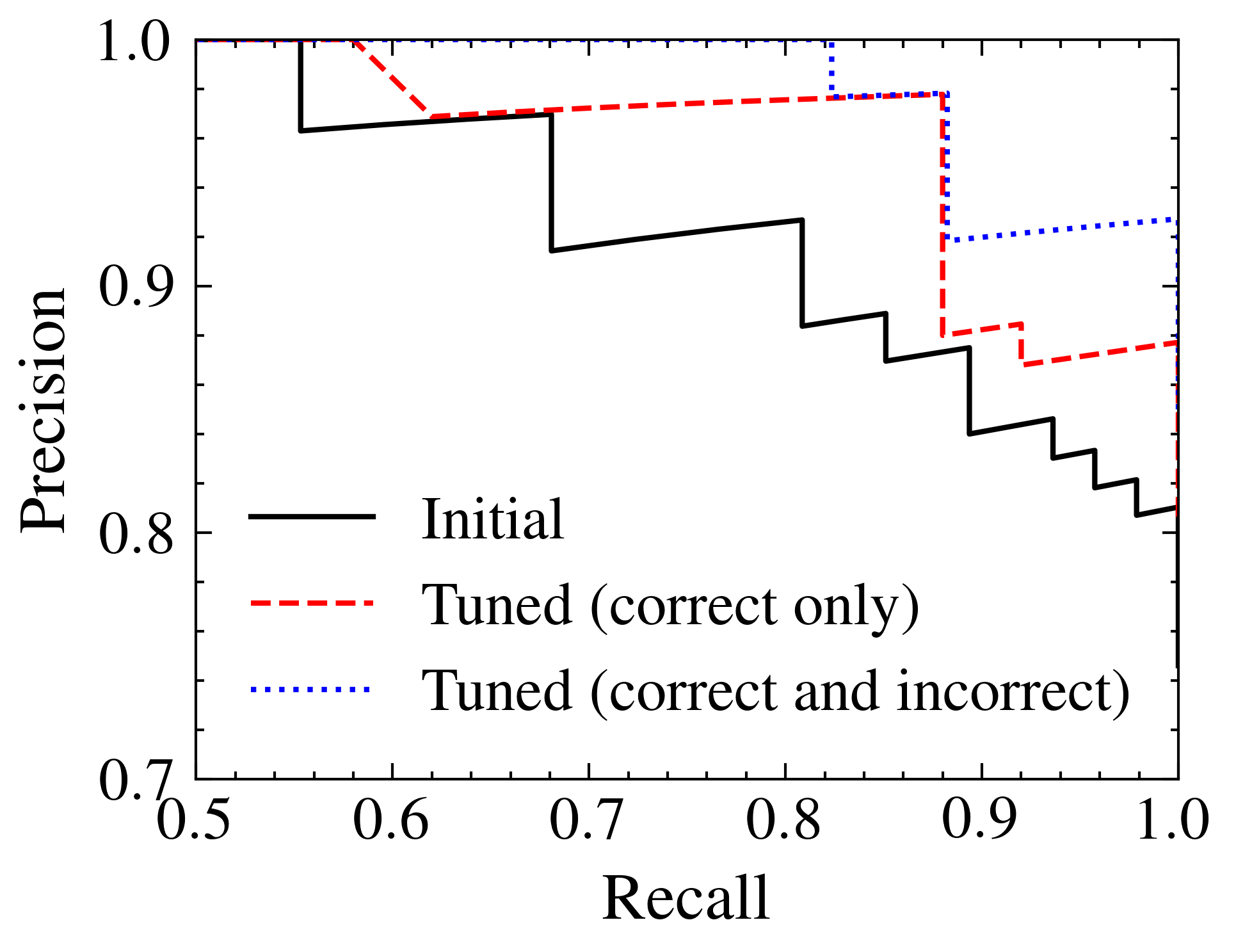}
        \caption{NetVLAD KITTI~06}
    \end{subfigure}

    \begin{subfigure}[b]{0.245\textwidth}
        \includegraphics[width=\textwidth,trim=0mm 0mm 0mm 0mm,clip]{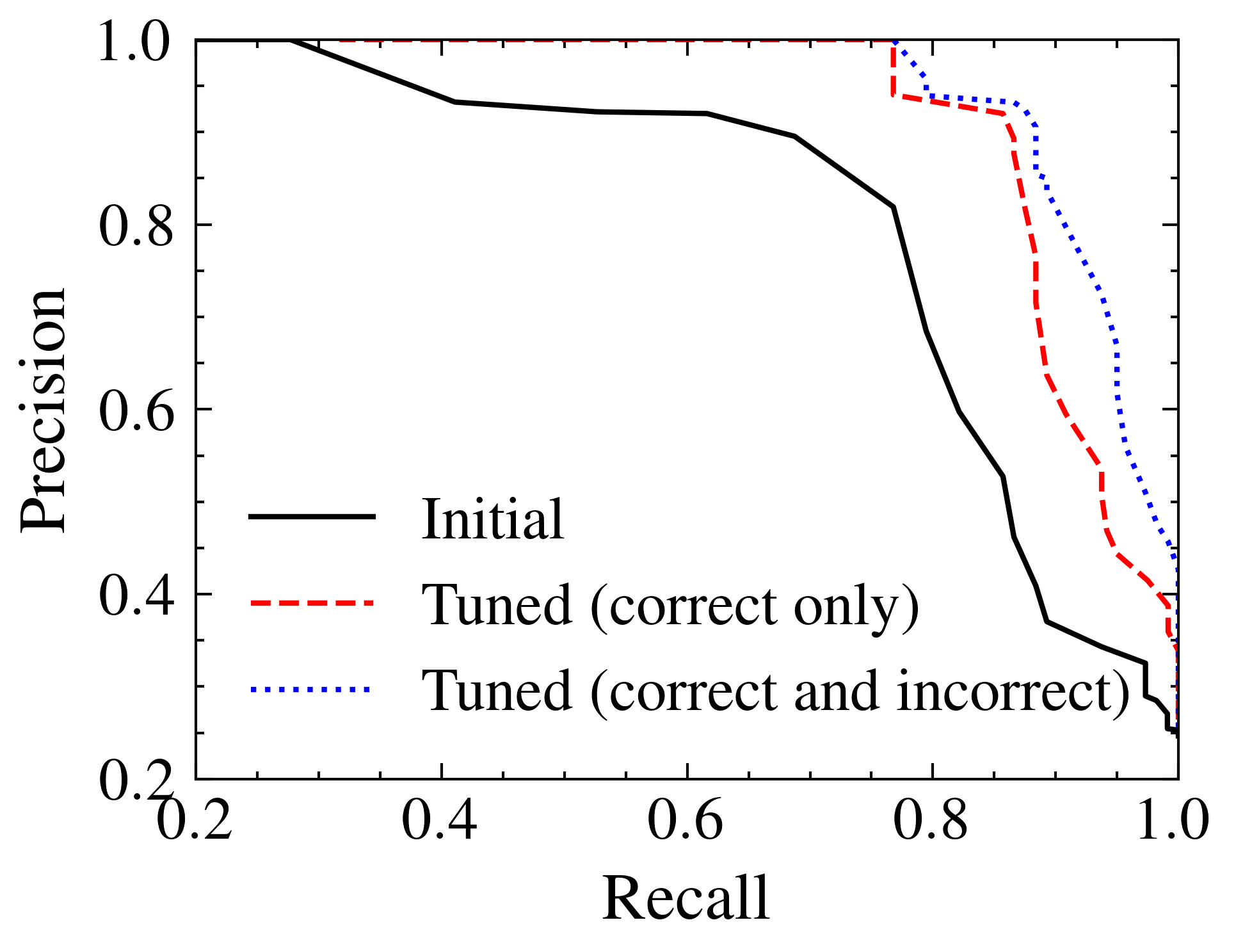}
        \caption{ViT KITTI~00}
    \end{subfigure}
    \begin{subfigure}[b]{0.245\textwidth}
        \includegraphics[width=\textwidth,trim=0mm 0mm 0mm 0mm,clip]{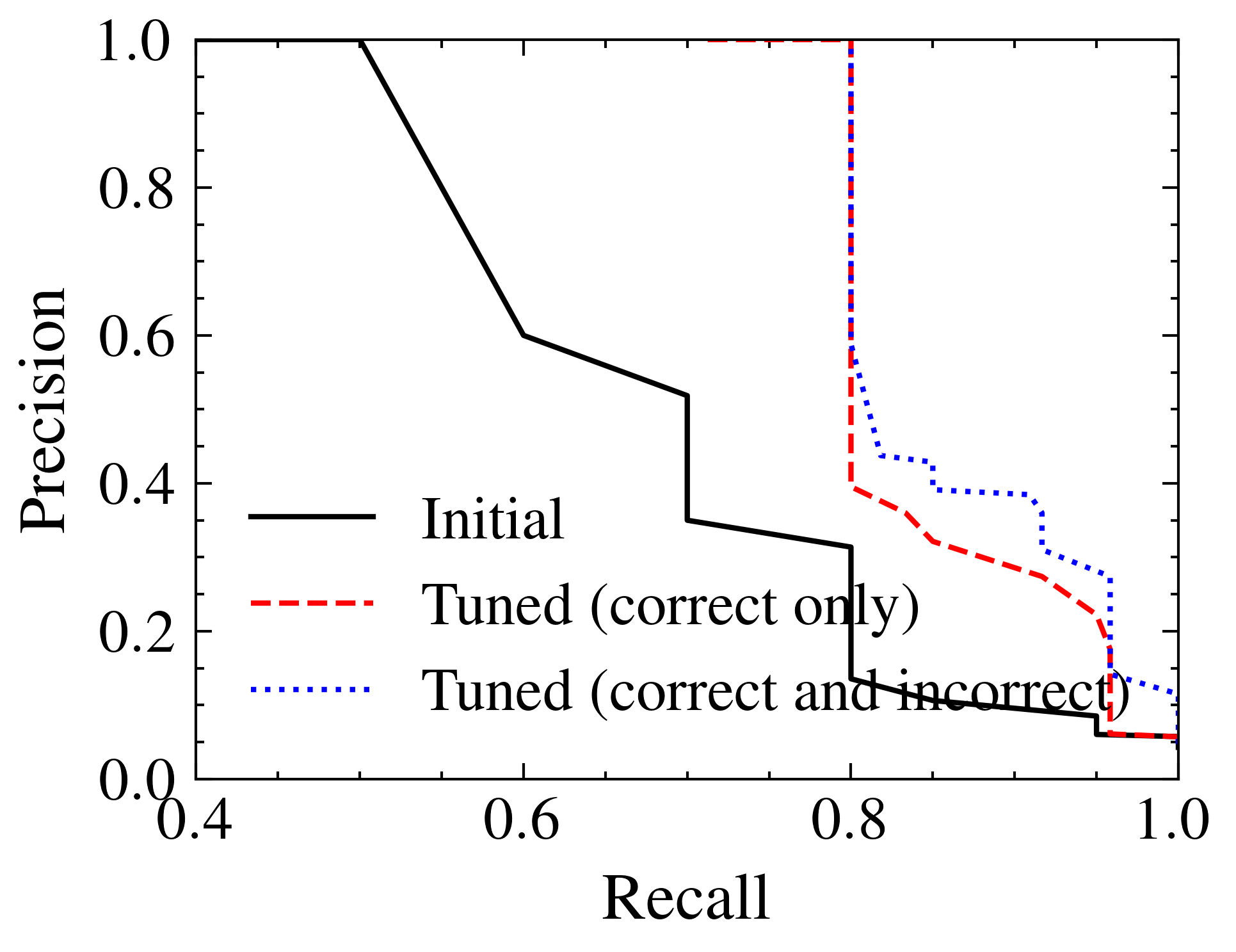}
        \caption{ViT KITTI~02}
    \end{subfigure}
    \begin{subfigure}[b]{0.245\textwidth}
        \includegraphics[width=\textwidth,trim=0mm 0mm 0mm 0mm,clip]{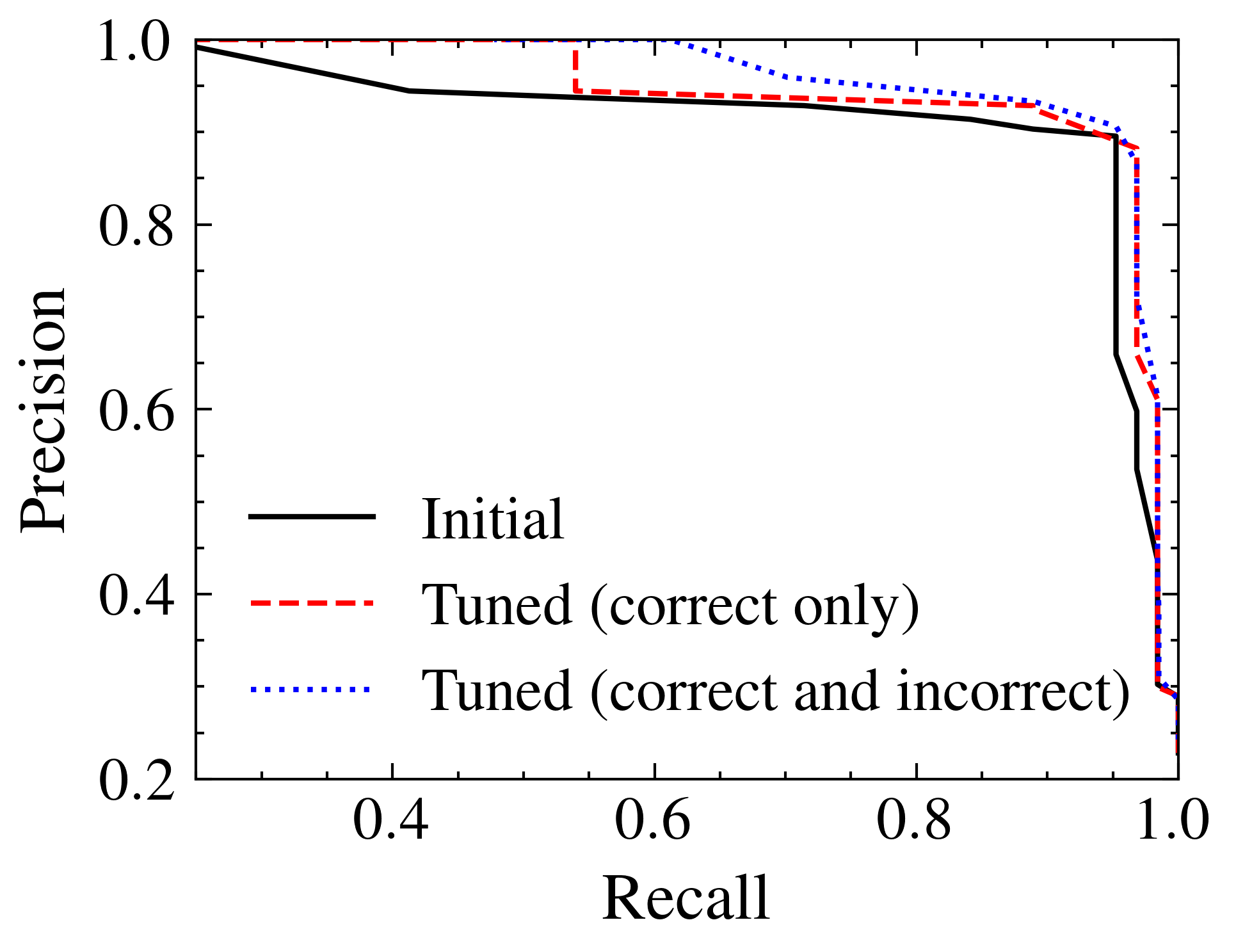}
        \caption{ViT KITTI~05}
    \end{subfigure}
    \begin{subfigure}[b]{0.245\textwidth}
        \includegraphics[width=\textwidth,trim=0mm 0mm 0mm 0mm,clip]{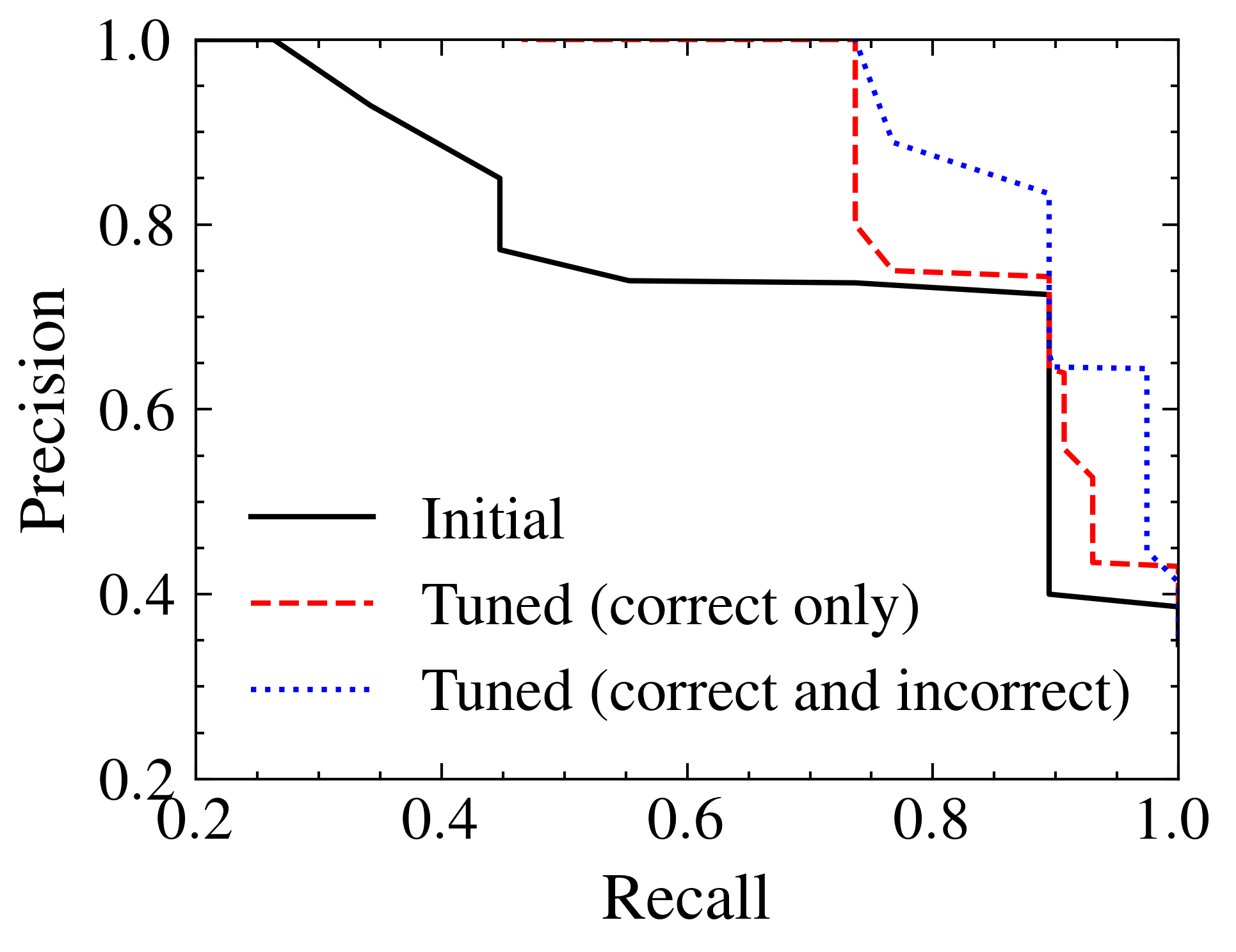}
        \caption{ViT KITTI~06}
    \end{subfigure}
    \caption{Precision-Recall performance in loop-closure detection on various KITTI sequences with two different network architecture: NetVLAD~\cite{arandjelovicNetVLADCNNArchitecture2018} and ViT~\cite{dosovitskiyImageWorth16x162021}. As expected, we can see that the networks tuned on the KITTI-360-09 sequence achieves better precision and recall on all sequences.}
    \label{fig:precision_recall}
\end{figure*}

\subsection{Uncertainty Estimation}
\label{subsec:uncertainty}

To learn an estimator tuned to the desired target environment, we
follow~\cite{warburgBayesianTripletLoss2021b} and add an uncertainty head to the
baseline CNN network composed of a Generalized Mean (GeM)
layer~\cite{radenovicFineTuningCNNImage2019} followed by two fully connected
layers with a softplus activation function. The resulting network has two
outputs, a mean $\mu$ and a variance $\sigma$, such that it encodes the
descriptors as isotropic Normal distributions $\mathcal{N}(\mu, \sigma)$ instead
of point estimates.

For each correct sample, the uncertainty-aware
 loss computes the probability that the query $q$ is closer to a positive $p$ than a
negative $n$ given a margin $m$, and a prior $1/K$ for normalization,
\begin{align}
\label{eq:probtriplet1}
    P\left(\|q - p\|^2 < \|q - n\|^2 - m\right),
\end{align}

For incorrect samples, we use instead the Kullback-Leibler divergence $D_{KL}(V
\| T)$ loss between the descriptors distribution estimates ($V \in
\mathcal{N}(\mu, \sigma)$) and a target high variance distribution with the same
mean ($T \in \mathcal{N}(\mu, \sigma_{high})$) for which we have set the
variance $\sigma_{high}$ to $H$ times the prior.
A higher $H$ increases the incorrect matches loss and thus their importance during training.
For isotropic Normal distributions, $D_{KL}(V\| T)$ is defined as follows~\cite{robertIntrinsicLosses1996},
\begin{align}
    D_{KL}(V \| T) &= \frac{1}{2}\left(\log \frac{\sigma^2_{high}}{\sigma^2} + \frac{\sigma^2}{\sigma_{high}^2} - 1\right)
\end{align}
We sum the Kullback-Leibler divergences of
the query $q$ ($D_{KL}(V_q \| T)$) and the negative $n^q$ ($D_{KL}(V_{n^q} \| T)$) to obtain the combined loss $L$ of the incorrect sample $s \in S_{incorrect}$,
\begin{align}
    L&= \frac{1}{2}\left(D_{KL}(V_q \| T) + D_{KL}(V_{n^q} \| T)\right)
\end{align}
The intuition behind the use of this loss function is to increase the variance
estimates of both the query and negative images towards a higher variance
without changing their means, since those images are confusing for the VPR
system and led to loop closing outliers.

It is important to note that training using uncertainty-aware losses can have
detrimental effects on the resulting precision of the place recognition
network~\cite{tahaUnsupervisedDataUncertainty2019}. However, as we show in
the following section, a variance estimator with reasonable performance can be
trained on a separate smaller network that can be run cheaply on a CPU. This
could allow practitioners to keep the precision of a conventionally trained VPR
network and run a smaller uncertainty estimation network in parallel.

\section{Experiments}
\label{sec:exp}

Our experiments are divided into three parts. We first demonstrate the quality
of the extracted training tuples by using them to train a network for the task
of place recognition from a classification baseline. Second, we demonstrate that
we can calibrate a state-of-the-art VPR approach for a target domain using our
technique. Third, we show that we can achieve uncertainty estimates
tailored to the target environment. All the hyperparameters values used in our
experiments can be found in our open-source implementation and correspond to the
ones used in~\cite{arandjelovicNetVLADCNNArchitecture2018}
and~\cite{warburgBayesianTripletLoss2021b}.

\subsection{Training a new VPR System}

To show the effectiveness of our approach to produce valuable tuning samples
from a calibration run through an environment, we trained a new VPR network
based exclusively on the samples extracted from a single
KITTI-360~\cite{liaoKITTI360NovelDataset2022} sequence and we tested the resulting
VPR network on KITTI~\cite{geigerAreWeReady2012} sequences. All the sequences
were collected in the streets of the same mid-size city. We used the
KITTI-360-09 sequence for tuning, and the KITTI-\{00, 02, 05, 06\}
sequences for
testing. The testing sequences were collected years apart from the tuning
sequence and were selected based on the significant overlaps within their
trajectory, which are essential to recognize places.

Our initial model consist of a VGG16 network pretrained on ImageNet~\cite{Simonyan15} for
which we replaced the classification head with a randomly initialized NetVLAD pooling
layer~\cite{arandjelovicNetVLADCNNArchitecture2018}.
To show the generalization of the approach on different network architectures, we also tuned a Vision Transformer (ViT)~\cite{dosovitskiyImageWorth16x162021} using the penultimate layer features as descriptors.

The new networks were tuned for place recognition using a triplet margin loss
 for 10 epochs, which was enough to achieve convergence.
The relative poses, loop closures, are estimated with stereo pairs and the SLAM visual odometry is
computed and managed using RTAB-Map~\cite{labbeRTABMapOpensourceLidar2019}. Our
technique successfully extracted 291 training tuples in $S_{correct}$, and 49 in $S_{incorrect}$, from the tuning sequence.

To validate the training procedure, we computed the similarity score, based on
the $L_2$ distance between global descriptors, of all pairs of images in
KITTI-00 sequence. In \cref{fig:distance_matrix_training_kitti}, we compared the
resulting similarity matrix with the one before tuning and the one obtained with
NetVLAD. NetVLAD, which is pretrained on city streets images, is known to
achieve high accuracy on the KITTI
sequences~\cite{cieslewskiDataEfficientDecentralizedVisual2018}. We can see that
our approach converges to a similar result as NetVLAD, especially in the zones
where multiples places are revisited and recognized (i.e., high similarity) near
the bottom left and right corners (darker blue). The contrast with negative
matches is also accentuated.

\begin{figure}[h]
    \centering
    \begin{subfigure}[b]{0.43\columnwidth}
        \includegraphics[width=\columnwidth,trim=0mm 0mm 0mm 0mm,clip]{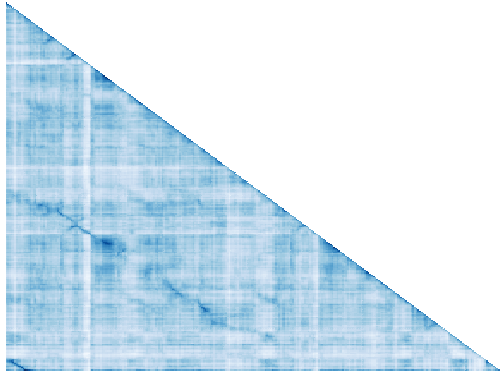}
        \caption{Original NetVLAD}
    \end{subfigure}
    \begin{subfigure}[b]{0.43\columnwidth}
        \includegraphics[width=\columnwidth,trim=0mm 0mm 0mm 0mm,clip]{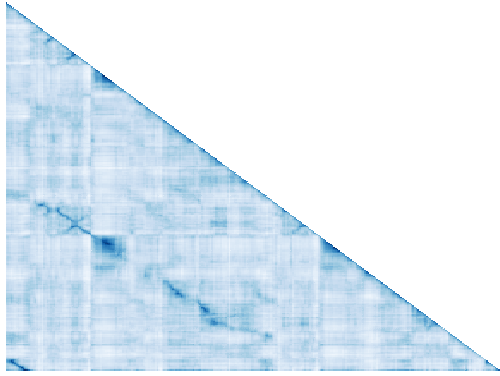}
        \caption{Tuned NetVLAD}
    \end{subfigure}
    \begin{subfigure}[b]{0.055\columnwidth}
        \includegraphics[width=\columnwidth,trim=0mm 0mm 0mm 0mm,clip]{figures/sim_bar.png}
        \caption*{}
    \end{subfigure}
    \caption{Self-Supervised domain calibration of a visual-similarity metric.
      An illustration of the similarity matrix before (Original NetVLAD) and
      after training (Tuned NetVLAD) compared. As expected, the similarity
      between positive pairs has increased (blue), and it has decreased between
      negative pairs (white).}
    \label{fig:distance_matrix_training_lab}
\end{figure}

\begin{figure}[h]
    \centering
    \begin{subfigure}[b]{0.4925\columnwidth}
        \includegraphics[width=\columnwidth,trim=0mm 0mm 0mm 0mm,clip]{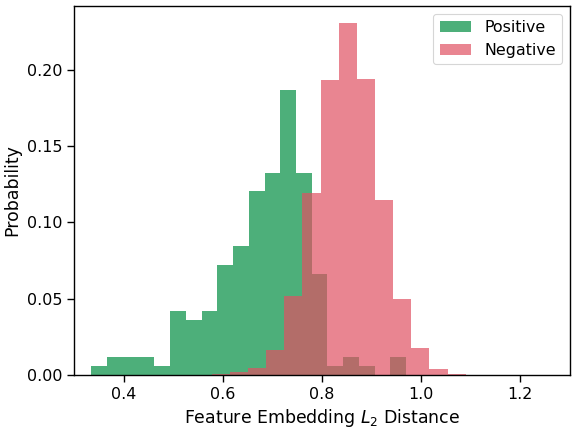}
        \caption{Original NetVLAD}
    \end{subfigure}
    \begin{subfigure}[b]{0.4925\columnwidth}
        \includegraphics[width=\columnwidth,trim=0mm 0mm 0mm 0mm,clip]{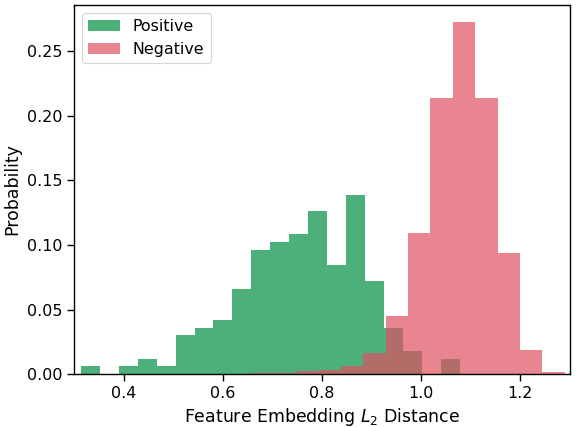}
        \caption{Tuned NetVLAD}
    \end{subfigure}
    \caption{Separation distance calibration. Histograms of the $L_2$ distance
      between the positive pairs (green) and negative pairs (red). As expected,
      the separation increased between the positive pairs and negative pairs
      after tuning, making it easier to set a VPR threshold.}
    \label{fig:embedding distance_lab}
    \vspace{-3mm}
\end{figure}

\begin{table}[t]
    \vspace{4mm}
    \centering
    \caption{Average percentage and standard deviation of correct matches obtained by NetVLAD before
      and after tuning. We can see that the domain calibration increased the
      percentage of correct matches and thus the number of loop closures.}
    \ra{1.2}
    \begin{tabular}{@{}lccccc@{}}\toprule
     & Indoor 1& \phantom{} & Indoor 2& \phantom{} & Indoor 3 \\
    \cmidrule{2-2} \cmidrule{4-4} \cmidrule{6-6}
    Original NetVLAD  &   $65.4\pm 8.9\%$ &&   $61.7\pm 12.4$ \% &&  $75.8\pm 5.9$ \% \\
    Tuned NetVLAD    &   $72.0\pm 8.7\%$ &&   $71.1\pm 10.6$ \%&&  $82.6\pm 4.8$ \% \\
    \bottomrule
    \end{tabular}
    \label{tab:metrics}
    \vspace{-5mm}
\end{table}

Using the ground truth poses of the KITTI sequences~\cite{geigerAreWeReady2012},
we computed the precision and recall of our VPR system resulting matches before
and after tuning, with or without the incorrect matches. The curves in
\cref{fig:precision_recall} represent the performance for varying detection
threshold values for loop closure detection. A loop closure is considered as
detected if the distance between the two images global descriptors is inferior
to the threshold, there exists sufficient keypoints matches to compute a
relative pose measurement, and it passes the test of robust pose graph
optimization. We can see a clear improvement in precision and
recall after tuning both NetVLAD and ViT. We also see some small improvements when using the incorrect matches,
especially on the KITTI~06 sequence, however we expect the effects of incorrect
matches to be greater for initial networks with low precision since it should reduce the number of false positives.
While state-of-the-art results were not expected on the well-studied KITTI
dataset, we are able to demonstrate the quality and efficiency of the gathered
training samples from a single run through the environment without
manual labelling or GPS bootstrapping.

\begin{figure*}[ht!]
    \centering
    \begin{subfigure}[b]{0.325\linewidth}
        \includegraphics[width=\linewidth,trim=0mm 0mm 0mm 0mm,clip]{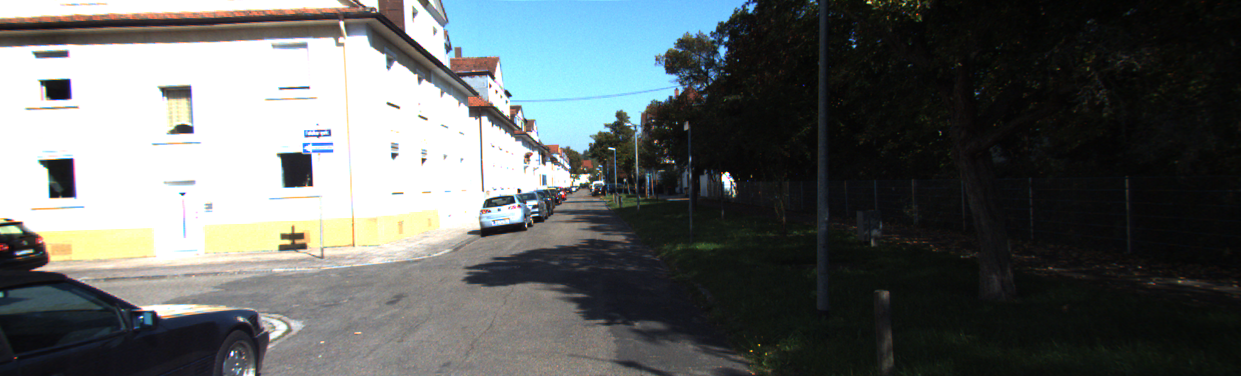}
    \end{subfigure}
    \begin{subfigure}[b]{0.325\linewidth}
        \includegraphics[width=\linewidth,trim=0mm 0mm 0mm 0mm,clip]{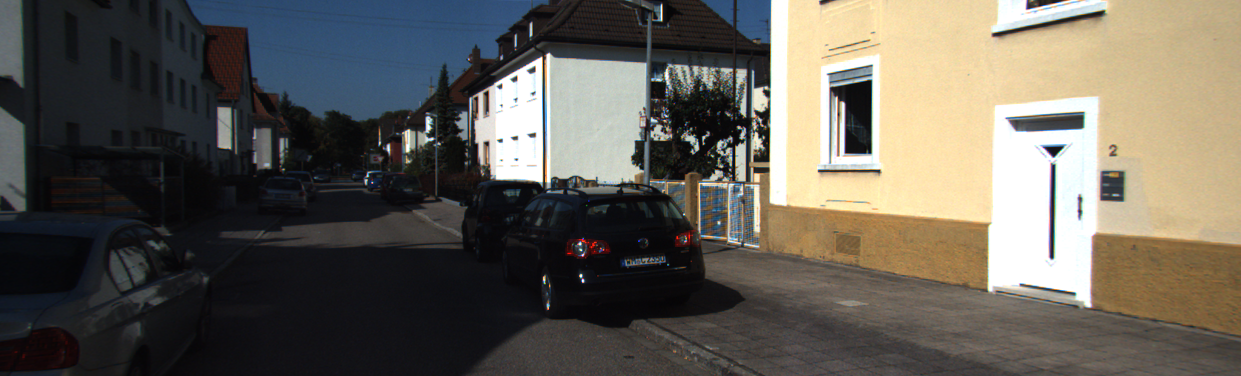}
    \end{subfigure}
    \begin{subfigure}[b]{0.325\linewidth}
        \includegraphics[width=\linewidth,trim=0mm 0mm 0mm 0mm,clip]{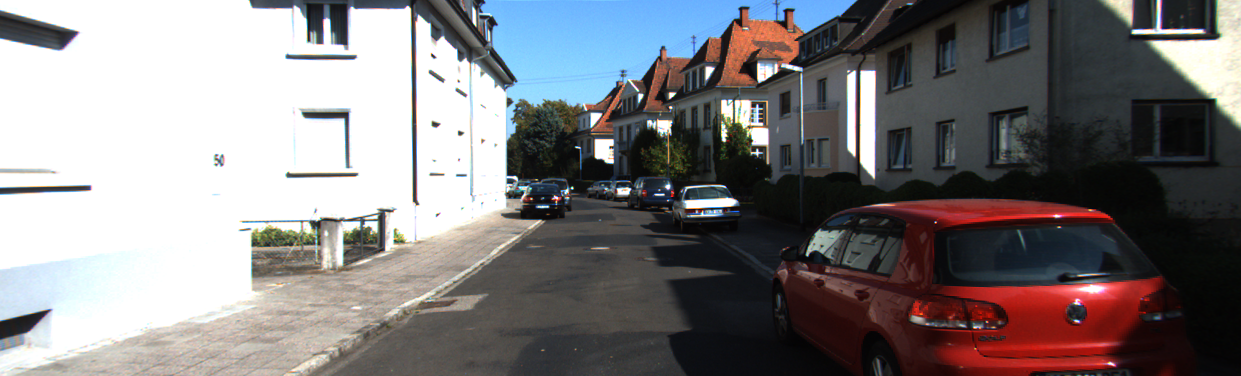}
    \end{subfigure}
    \caption{Examples of images with high estimated uncertainty. We can see the
      presence of under/overexposure and occlusions.}
    \label{fig:uncertainty_images}
\end{figure*}

\begin{table*}
    \centering
    \ra{1.2}
    \caption{Uncertainty Estimation results on KITTI sequences. We report the $F_1$ score (computed from precision and recall), the Expected Calibration Error using the cosine similarity between the descriptors as uncertainty estimates (${ECE}_{\text{sim}}$), and the $ECE$ using our trained uncertainty estimates (${ECE}_{\text{ours}}$) for each sequence with 2 different CNN backbone networks. We also indicate the networks size as well as their respective inference time on GPU (NVIDIA RTX3070) and CPU (AMD Ryzen 7).}
    \resizebox{\textwidth}{!}{
    \begin{tabular}{@{}lccc|ccc|ccc|ccc|ccc@{}}\toprule
    & \multicolumn{3}{c}{KITTI~00} & \multicolumn{3}{c}{KITTI~02} 
     & \multicolumn{3}{c}{KITTI~05}
     & \multicolumn{3}{c}{KITTI~06} &  \multicolumn{3}{c}{Resources}\\
    \cmidrule{2-4} \cmidrule{5-7} \cmidrule{8-10}  \cmidrule{11-13}  \cmidrule{14-16}
    & $F_1$ score & ${ECE}_{\text{sim}}$ & ${ECE}_{\text{ours}}$ & $F_1$ score  & ${ECE}_{\text{sim}}$ & ${ECE}_{\text{ours}}$& $F_1$ score  & ${ECE}_{\text{sim}}$ & ${ECE}_{\text{ours}}$ & $F_1$ score  & ${ECE}_{\text{sim}}$ & ${ECE}_{\text{ours}}$ & Size (MB) & GPU~(s) & CPU~(s)   \\ \midrule
    VGG16 & 0.848 & 0.774 & \textbf{0.186} & 0.815 & 0.801 & \textbf{0.226} & 0.821 & 0.769 & \textbf{0.125} & 0.732 & 0.802 & \textbf{0.305} & 82.0 & 0.004 & 0.052\\
    MobileNetv3 & 0.874& 0.949 & \textbf{0.242}& 0.704& 0.946 & \textbf{0.150} & 0.779& 0.948 & \textbf{0.197} & 0.720 & 0.945 & \textbf{0.367} & 23.8 & 0.007 & 0.015\\
    \bottomrule
    \end{tabular}
    }
    \label{tab:ece}
    \vspace{-3mm}
\end{table*}

\subsection{Calibration of an Existing VPR system}

In this set of experiments, we demonstrate that we can improve the performance
of a pretrained state-of-the-art VPR network
(NetVLAD~\cite{arandjelovicNetVLADCNNArchitecture2018}) by tuning it to a
different target domain. We performed four runs through an indoor office
environment (see images in \cref{fig:top-level}) using an Intel Realsense D455
camera, with multiple overlaps to ensure place recognition. The first run served
to extract 166 training tuples, and the three others have been used for testing.

As shown in \cref{fig:distance_matrix_training_lab}, we computed the similarity
score for each image pairs in the training sequence before (i.e., original pretrained
version of NetVLAD on the Pittsburg dataset~\cite{arandjelovicNetVLADCNNArchitecture2018}) and after tuning. 
We can observe a significant improvement
in contrast between positive and negative matches hinting to a better
distinction between them during testing.

We corroborate this result in~\cref{fig:embedding distance_lab} which shows
histograms of the $L_2$ distance between all pairs of images. The positive pairs
are noted in green and the negative ones in red. Confirming the previous result,
the separation between the positive and negative pairs is greater after tuning.
\cref{fig:embedding distance_lab} shows that our calibration technique is able
to fine-tune the VPR network and distort the feature embedding to increase the
distance between similar and dissimilar places. This has practical implications
for the deployment of VPR systems since the threshold to determine if the
distance represents a match becomes easier to set. Moreover, our approach leads to
fewer false positives which can be detrimental to the system accuracy and
computational performance~\cite{carsonPredictingImproveIntegrity2022}.
In \cref{tab:metrics} we confirm on the three test sequences that NetVLAD tuned
with our technique obtains on average a significantly higher number of correct VPR matches
over possible threshold values than its original version (t-test, Bonferroni-corrected, $p <$ 1e-5). Therefore, our approach allows
practitioners to improve the performance of their VPR system by calibrating its
domain through a single run of the environment.

\subsection{Uncertainty Estimation for VPR}

As expected, results in~\cref{tab:ece} show that training a network explicitly for uncertainty estimation performs better than directly using the cosine similarity between descriptors as a confidence measure.
Also, we noticed that training with an uncertainty loss
does not provide as much improvement in precision and recall as the
triplet margin loss. However, having uncertainty estimates is
preferable in safety critical applications. One could even combine a network
trained for state-of-the-art precision and another trained for uncertainty in
the same system if the computing resources are sufficient. To that end, we show
that we can achieve reasonable results in uncertainty estimation with a smaller
backbone network (MobileNetv3~\cite{howardSearchingMobileNetV32019}) which can
be run in real-time on a CPU. This decoupling offers flexibility for practical deployments of uncertainty-aware visual place
recognition.

In \cref{fig:uncertainty_images}, we present some examples of images with high
estimated uncertainties by our technique. The uncertain images are
under/overexposed and have occlusions, such that very few useful visual features
and keypoints could be extracted from them in order to successfully compute 3D
registration.

To measure the accuracy of the uncertainty estimates, we use the
\textit{Expected Calibration Error} (ECE), commonly used for classification
tasks~\cite{gustafssonEvaluatingScalableBayesian2020}, which computes how well
the uncertainty estimates correspond to the model's precision,
\begin{align}
    \mathrm{ECE}=\sum_{m=1}^{M} \frac{\left|B_{m}\right|}{n}\left|\operatorname{mAP@1}\left(B_{m}\right)-\operatorname{conf}\left(B_{m}\right)\right|.
\end{align}  

As in~\cite{caiSTUNSelfTeachingUncertainty2022a}, we compute this metric by dividing the uncertainty estimates into $M$ equally
spaced bins $B_{m}$ with corresponding uncertainty level $U(B_{m})$. 
For each bin, we compute the precision of the queries it contains, $\operatorname{mAP@1}\left(B_{m}\right)$, and compare it with the bin confidence
$\operatorname{conf}\left(B_{m}\right) = 1-U(B_{m})$. 
The ECE is low when the high confidence
images lead to high precision matches. As expected, the resulting ECEs in
\cref{tab:ece} are in a similar range as the results
presented in the Bayesian Triplet Loss
paper~\cite{warburgBayesianTripletLoss2021b}. Interestingly for
resource-constrained deployments, the smaller MobileNetv3 achieves comparable
results to the larger VGG16 while being able to evaluate images in real-time at
more than 60Hz on a CPU.
\section{Conclusions}
\label{sec:conclusion}

In this paper, we present a self-supervised method for training and tuning a
place recognition neural network leveraging robust SLAM which does not require
GPS or ground truth labels for bootstrapping. We demonstrate the efficiency of
the method by training a visual place recognition network from a pretrained
classification model, using only the training samples extracted by our method.
We also show that our technique can improve the accuracy of an existing deep
learning-based VPR system by calibrating it to the target environment. In
addition, we show that we can train an uncertainty estimation network for place
recognition using the extracted samples.

We consider that our approach has practical benefits for the real-world
deployment of place recognition systems. It could be used in an online fashion
to perform lifelong learning/tuning on the target environment. Our approach has
also the potential for data mining of labeled place recognition training samples
on any sequential dataset, which could help increased the overall accuracy of
VPR networks. Moreover, while we applied our technique to visual sensors, the
same approach could be used for other type of sensors used for place recognition
(e.g. lidars).
Finally, we believe that leveraging the recent progress in robust SLAM to improve the performance of deep learning based techniques is a
promising avenue that could lead to a tighter integration between the two fields
of research.

\bibliographystyle{IEEEtran}

\end{document}